\documentclass[letterpaper, 10 pt, conference]{ieeeconf}  

\IEEEoverridecommandlockouts                              

\usepackage{xcolor}
\usepackage[pdftex]{graphicx}
\usepackage{subfigure}
\usepackage{makecell}
\usepackage{amssymb}
\usepackage{amsmath}
\usepackage{bm}
\usepackage[pagebackref,breaklinks,colorlinks,citecolor=blue]{hyperref}
\usepackage{makecell}
\usepackage{xcolor}
\usepackage[linesnumbered,ruled,vlined]{algorithm2e}
\usepackage{algpseudocode}
\usepackage{subfiles}
\usepackage{siunitx}

\DeclareMathOperator{\trace}{tr}

\DeclareMathAlphabet\mathbfcal{OMS}{cmsy}{b}{n}

\usepackage[textsize=footnotesize]{todonotes}

\title{\LARGE \bf
MonoForce: Self-supervised Learning of Physics-informed Model for Predicting Robot-terrain Interaction
}

\author{Ruslan Agishev$^{1, 2}$, Karel Zimmermann$^{1}$, Vladim{\' i}r Kubelka$^{2}$, Martin Pecka$^{1}$ and Tom{\' a}{\v s} Svoboda$^{1}$
\thanks{$^{1}$The authors are with the VRAS group, Faculty of Electrical Engineering, Czech Technical University in Prague, Czech Republic (e-mail: agishrus@fel.cvut.cz; zimmerk@fel.cvut.cz; peckama2@fel.cvut.cz; svobodat@fel.cvut.cz)
\emph{(Corresponding author: R.~Agishev.)}
}
\thanks{$^{2}$The authors are with the RNP Lab of the AASS Research Centre, Örebro University, Örebro, Sweden (e-mail: vladimir.kubelka@oru.se; agishrus@fel.cvut.cz)}
}

\begin{document}

\maketitle
\thispagestyle{empty}
\pagestyle{empty}

\begin{abstract}
While autonomous navigation of mobile robots on rigid terrain is a well-explored problem, navigating on deformable terrain such as tall grass or bushes remains a challenge.
To address it, we introduce an explainable, physics-aware and end-to-end differentiable model which predicts the outcome of robot-terrain interaction from camera images, both on rigid and non-rigid terrain. 
The proposed MonoForce model consists of a black-box module which predicts robot-terrain interaction forces from onboard cameras, followed by a white-box module, which transforms these forces and a control signals into predicted trajectories, using only the laws of classical mechanics.
The differentiable white-box module allows backpropagating the predicted trajectory errors into the black-box module, serving as a self-supervised loss that measures consistency between the predicted forces and ground-truth trajectories of the robot.
Experimental evaluation on a public dataset and our data has shown that while the prediction capabilities are comparable to state-of-the-art algorithms on rigid terrain, MonoForce shows superior accuracy on non-rigid terrain such as tall grass or bushes.
To facilitate the reproducibility of our results, we release both the code and datasets.

\end{abstract}

\section{INTRODUCTION}
The ability to predict the outcome of robot-terrain interaction from onboard sensors is essential for many fundamental functionalities of autonomous navigation, such as traversability estimation, planning and control.
Over the last few decades, a wide variety of \emph{white-box}~\cite{Fabian2020, Dogru-AuRo-2021, manoharan-IROS-2024} and \emph{black-box}~\cite{Wellhausen-RAL-2019, hdif2023, li2023seeing} motion models have been proposed, each with its strengths and weaknesses. 
We propose to get the best of both worlds by introducing a \emph{grey-box}, explainable, physics-aware, and end-to-end differentiable model that enables self-supervised learning.

\begin{figure}[t]
\centering
\includegraphics[width=0.48\textwidth]{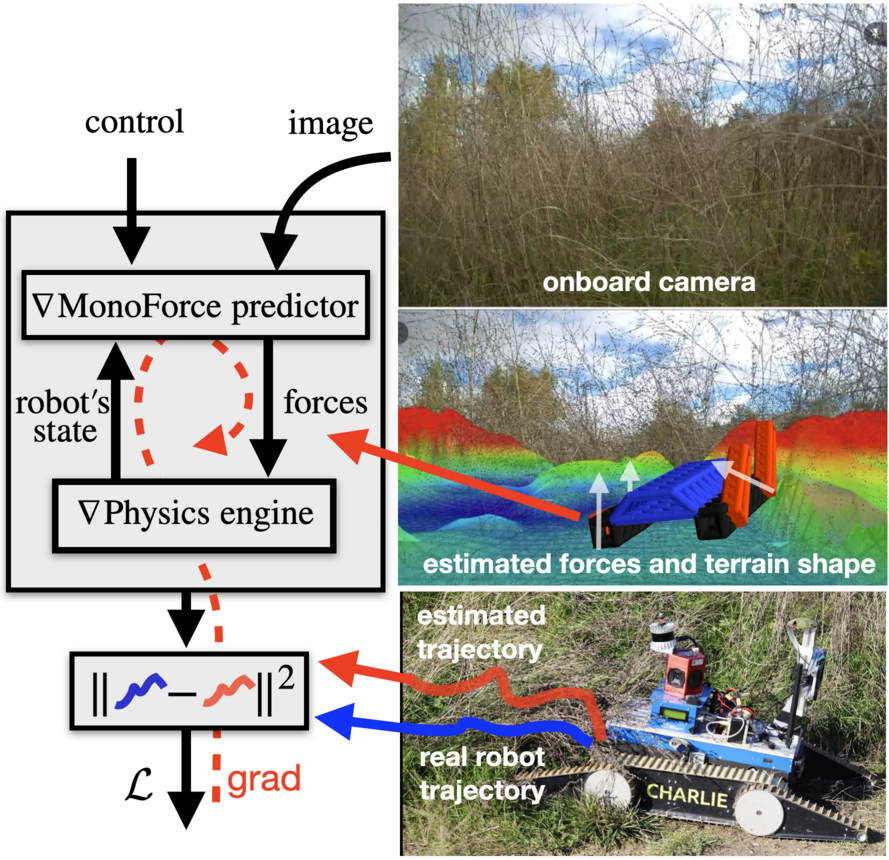}
  \caption{\textbf{Architecture overview:} \emph{$\nabla$MonoForce predictor} delivers robot-terrain reaction forces conditioned by an image from the onboard monocular camera, by the robot state, and by the control input. \emph{$\nabla$Physics engine} integrates these forces to estimate the resulting robot trajectory. 
  Self-supervised loss measures physical consistency between forces and the real robot trajectory through the differentiable physics engine.
  }
  \label{fig:overview}
\end{figure}

We focus on predicting the robot's trajectory on both rigid and deformable terrain from a single image captured by an onboard camera. 
The proposed model is based on a physics-aware simulation conditioned by the terrain image, see Figure~\ref{fig:overview} for details.
The model consists of a \emph{MonoForce predictor}, which delivers robot-terrain reaction forces given the monocular image, state and control, and a \emph{physics engine}, which integrates these forces and estimates the resulting robot's trajectory. While the physics engine is mostly deterministic, the transformation from image to reaction forces is unknown. A~natural question emerges: ``How can we train the MonoForce predictor?''.



It has been shown that supervised learning of robot-terrain reaction forces is possible for legged robots~\cite{li2023seeing} because the force appears only on its small-sized feet that are in contact with the terrain. 
Such force is inherently well-localized in the camera frame, and its magnitude and direction can be measured accurately through torque sensors.
Therefore, the training data can be easily provided. 
However, direct measurement of contact forces for wheeled or tracked robots~\cite{Inoue2008} requires highly specialized sensors and is typically hard to achieve in outdoor scenarios.
Thus, the contact forces have to be inferred from robot trajectories.  
Since both the MonoForce predictor and the physics engine are fully differentiable in our implementation, we train the predictor by backpropagating the trajectory error through the physics engine into the predictor parameters. 
In such a setup, the physics engine plays the role of a self-supervised loss that measures the consistency between predicted forces and ground truth trajectories.
The resulting setup then resembles the well-known MonoDepth model~\cite{monodepth2}, which learns to predict depth from the image by optimizing \emph{color consistency} between the predicted depth and the image delivered by another calibrated camera onboard.


The most straightforward architecture of the motion model would be a black-box, deep-convolutional network that would transform images and control commands into a 3D force field over the state space.
However, since there is an inherent training/testing distribution mismatch due to the natural absence of robot-endangering samples in the real training data, a good generalization of the force predictor is crucial.
Consequently, we search for architectures that combine black-box deep convolution layers with the white-box laws of classical mechanics and camera geometry.
The resulting model builds on the Lift-Splat-Shoot architecture~\cite{philion2020lift} that predicts the terrain properties, such as the shape of the terrain or its rigidity, from a single image and delivers forces that are directly computed from these properties. 
We emphasize that the resulting model is also directly differentiable with respect to the robot model. Therefore, joint optimization of the robot model, such as its center of gravity or moment of inertia, is also possible. The model is also directly differentiable with respect to control; therefore, its plug-and-play application in state-of-the-art MPC controllers~\cite{Amos-NEURIPS-2018} is at hand.
\textbf{Our main contributions} are as follows:

\textbf{Grey-box model:} A novel end-to-end differentiable grey-box model for predicting robot behavior on complex, non-rigid terrains from a monocular image.

\textbf{Self-supervised learning:} A physics-aware, self-supervised learning setup that delivers robot-terrain reaction forces and terrain model for tracked and wheeled robots with no contact force or torque sensors.

\textbf{Experimental evaluation on non-rigid terrains:} The proposed model outperforms other state-of-the-art methods on non-rigid terrains, such as grass or soft undergrowth that deforms when traversed by the robot.

See the project repository\footnote{\url{https://github.com/ctu-vras/monoforce}} for codes, dataset and supplementary materials such as videos and presentations.

\begin{figure*}[th]
\centering
\includegraphics[width=0.99\textwidth]{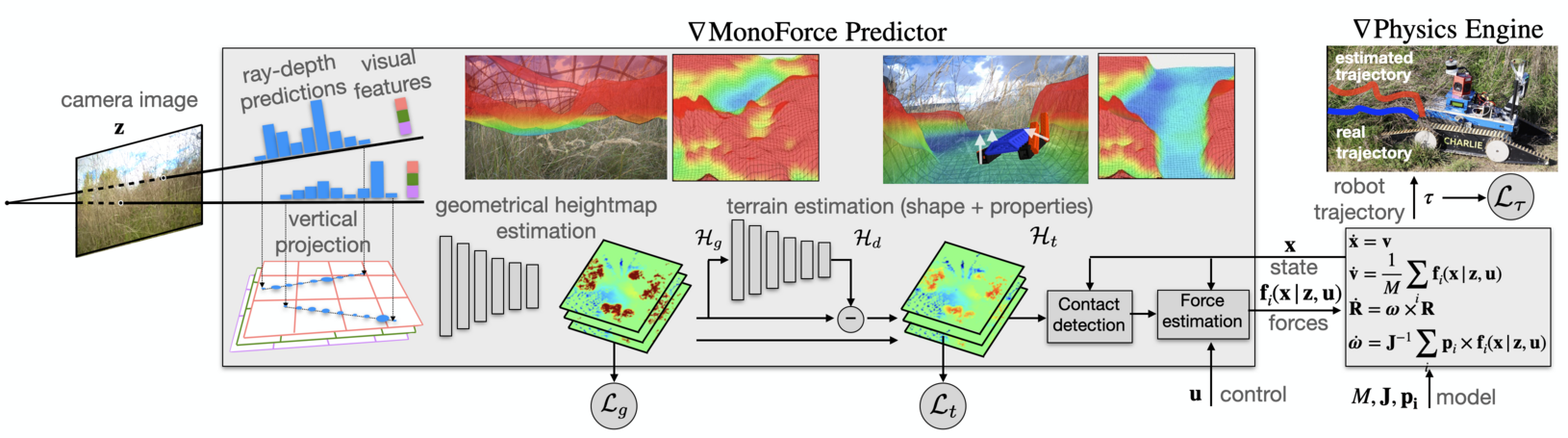}
  \caption{\textbf{Detailed architecture overview of "$\nabla$MonoForce predictor" and "$\nabla$Physics engine":} Neural network estimates depth predictions and rich visual features for each pixel ray. Depth-weighted visual features are vertically projected on 2.5D representation, and a geometrical heightmap is estimated. This heightmap is further refined through terrain predictor. It delivers terrain properties such as the heights of the rigid layer of terrain hidden under the vegetation or stiffness and dampening. Given state, control and terrain properties, forces at robot-terrain contacts are estimated. Finally, \emph{$\nabla$Physics engine} integrates these forces to estimate the resulting robot trajectory. We employ two losses: Trajectory loss, which measures the distance between the predicted and real trajectory, and an auxiliary geometrical loss, which measures the distance between the predicted geometrical heightmap and lidar-estimated heightmap.
  }
  \label{fig:architecture}
\end{figure*}
\section{RELATED WORK}
Despite being extremely important for predicting robot behavior on complex terrains, most of the so far introduced motion models fall into one of two categories: white-box or black-box models. We argue that this is one of the most substantial reasons for failure in outdoor robotics. This work is a~step towards filling the grey-zone gap in order to combine the strengths of both types of models.

White-box models often provide good generalizations if prior assumptions, such as terrain rigidity, are satisfied.
For example, Fabian~et~al.~\cite{Fabian2020} introduce the concept of robot heightmap and a novel iterative geometric method that allows fast 6DOF pose estimation on terrain heightmaps.
For skid-steered wheeled robots, Dogru and Marques~\cite{Dogru-AuRo-2021} propose a kinematic model that takes into account both the geometry and weight distribution of the robot.
They test their approach on solid surfaces with different slip properties.
However, the unaccounted and hard-to-model influence of non-rigid terrain, which is typically present in vegetation-rich environments, limits the applicability of these white-box models.

Black-box models can be trained on an arbitrary domain; however, the training data are restricted to robot-non-threatening samples.
For example, Wellhausen et~al.~\cite{Wellhausen-RAL-2019} project foothold positions of a walking quadruped robot into the images of its on-board camera.
Then, this data is used to train a~convolutional network to predict terrain properties relevant for the walking robot. 
Castro et al.~\cite{hdif2023} propose a method to learn a prorioceptive-based traversability cost from exteroceptive visual and geometrical data, in a self-supervised manner.
They demonstrate the method on two different all-terrain robots and it leads to smoother navigation.
The lack of extreme training examples (e.g. robot falling or rolling over) put together with poor generalization outside the training domain lead to the fact that black-box models usually do not perform well in corner-case situations.

This issue can be worked around by providing manual annotations for real data. Palazzo et~al. combine manually annotated real data with a simulator using domain transfer~\cite{Palazzo-IROS-2020}. However, manual annotations for traversability are extremely expensive and error-prone. A~cheaper source of manual annotations are expert driver trajectories, that can be used to learn to predict costmap using Inverse Reinforcement Learning~\cite{Silver-IJRR-2010}. Although IRL can theoretically learn from robot-endangering samples (since the driver will avoid them), it is often ill-conditioned and the assumption that the driver provides optimal trajectories with respect to a~consistent latent cost is often violated.

Explainable (grey-box) models also start appearing in the robotics literature. Zeng~et~al.~\cite{Urtasun2019} construct a DNN with partially explainable internal structure and guide the network to learn not only plans for driving an autonomous car, but also the intermediate products -- detecting other cars, object bounding boxes and motion vectors. However, this method needs lots of manually annotated data and externally provided maps to train the explainable inner parts correctly. We argue that self-supervised methods are more suitable for outdoor driving where it is much more problematic to provide high-quality manual annotations or semantic maps.

Even though manually supervised approaches will always be superior to self-supervised methods (if sufficient amount and quality of manually annotated data is provided), we do not consider them to be a direct competitor to the proposed method because manual annotations do not scale well and cannot be easily transferred among different domains (environments and robots).

Finally, the proposed $\nabla$physics module is implemented as a differentiable ODE solver. The existing end-to-end differentiable physics solvers include Neural ODE framework~\cite{neural-ode-2021} or differentiable simulators such as Google’s brax~\cite{brax2021} or Nvidia’s WARP~\cite{warp2022}. While the feedforward pass is usually based on Runge-Kutta integrator with a dynamic temporal step, backpropagation through physics is typically tackled by the implicit function theorem. We found out that for our task, the custom-built Euler integrator with fixed temporal step delivers similar accuracy while being approximately $10\times$ faster.


\section{THEORY}

A detailed overview of the proposed architecture, which converts images and control commands into robot trajectories, is shown in Figure~\ref{fig:architecture}. 
Given an input image $\mathbf{z}$ and robot control $\mathbf{u}$, the proposed architecture successively estimates geometric heightmap $\mathcal{H}_g$, terrain heightmap $\mathcal{H}_t$, robot-terrain forces $\mathbf{f}_i$ and trajectories~$\tau$. The geometrical map $\mathcal{H}_g$ is the multichannel 2D array, which contains the heights of the environment observed in the camera in the first channel and visual features in the remaining channels (see Section~\ref{subsec:terrain_encoder} for details). Similarly, the terrain map $\mathcal{H}_t$ is the multichannel 2D array, which contains the heights at which the terrain is assumed to start generating forces against the robot in the first channel and visual features in the rest. The intuition is that it corresponds to a not fully flexible layer of terrain (e.g. mud) that is hidden under the fully flexible layer (e.g. small vegetation). It is estimated by subtracting the estimated heightmap decrease $\mathcal{H}_d$ from the geometrical heightmap. Both maps are augmented by visual features that are successively converted into terrain properties. The part of the architecture that predicts terrain properties is called the \emph{terrain encoder} (see Section~\ref{subsec:terrain_encoder}) for details). 
Given the predicted terrain properties, state of the robot and control commands (e.g. track speed or flipper position), the forces $\mathbf{f}_i$ acting on the robot are computed (see Sections~\ref{subsec:normal_forces} and~\ref{subsec:tangential_forces}). Finally, the physics engine solves the robot motion equation and estimates the trajectory corresponding to the delivered forces (see Section~\ref{subsec:physics_engine}). The learning process of the proposed architecture is detailed in Section~\ref{subsec:learning_and_lossses}.

\subsection{Terrain encoder}\label{subsec:terrain_encoder}

The proposed architecture starts by converting pixels from a 2D image plane into a heightmap with visual features. Since the camera is calibrated, there is a substantial geometrical prior that connects heighmap cells with the pixels. We incorporate the geometry through the Lift-Splat-Shoot architecture~\cite{philion2020lift}. This architecture uses known camera intrinsic parameters in order to estimate rays corresponding to particular pixels -- pixel-rays, Figure~\ref{fig:architecture}. For each pixel-ray, the convolutional network then predicts depth probabilities (blue histogram) and visual features (colored vector). Visual features are vertically projected on a virtual heightmap for all possible depths along the corresponding ray. The depth-weighted sum of visual features over each heighmap cell is computed, and resulting multi-channel array is further refined by deep convolutional network in order to estimate the terrain heightmap $\mathcal{H}_{t}$. 
We call this architecture \emph{terrain encoder}. 
The terrain encoder can be described as the following function:
$\mathcal{H}_{t} = h_{\mathbf{w}}(\mathbf{z})$,
where $\mathbf{w}$ is a vector of the model weights.
\subsection{$\nabla$Physics engine}~\label{subsec:physics_engine}
The higher the complexity of the physical model, the more complicated physical interactions can be modeled, and the lower inductive bias is consequently introduced. However, too complex models require detailed input parameters, such as the complete topological structure of the vegetation in front of the robot, which cannot be reconstructed from a single camera image. Learning to predict unobservable properties is prone to overfitting. To find a good trade-off between the inductive bias and the overfitting, we introduce a collection of physics models with increasing complexity. 

We model the robot as a rigid body represented by a~set of mass points $\mathcal{P} = \{(\mathbf{p}_i, m_i)\; | \; \mathbf{p}_i\in\mathbb{R}^3, m_i\in\mathbb{R}^+, i=1\dots N\}$, where $\mathbf{p}_i$ denotes coordinates of the $i$-th 3D point and $m_i$ its mass. We employ common 6DOF dynamics of a rigid body~\cite{contact_dynamics-2018} are as follows:
\begin{equation}
  \begin{split}
    \dot{\mathbf{x}} &= \mathbf{v}\\
    \dot{\mathbf{v}} &= \frac{1}{M}\sum_i\mathbf{f}_i
  \end{split}
  \quad\quad
  \begin{split}
    \dot{\mathbf{R}} &= \boldsymbol{\omega}\times\mathbf{R}\\
    \dot{\boldsymbol{\omega}} &= \mathbf{J}^{-1}\sum_i \mathbf{p}_i\times\mathbf{f}_i 
  \end{split}
  \label{eq:1-4}
\end{equation}
where $\mathbf{x}, \mathbf{v}\in\mathbb{R}^3$ are its position and velocity, $\mathbf{R}\in\mathcal{SO}_3$ is its rotation represented by $3\times 3$ orthogonal matrix and $\boldsymbol{\omega}\in\mathbb{R}^3$ is its angular velocity. We denote $\mathbf{f}_i\in\mathbb{R}^3$ all external forces acting on $i$-th mass point. Although the forces and the trajectory are assumed to be a function of time, we drop the time index for simplicity. Total mass $M=\sum_i m_i\in\mathbb{R}^+$ and moment of inertia $\mathbf{J}\in\mathbb{R}^{3\times 3}$ are assumed to be known static parameters of the rigid body since they can be identified independently in laboratory conditions. Note that the proposed framework allows backpropagating the gradient with respect to these quantities, too, which makes them jointly learnable with the rest of the architecture. The trajectory of the rigid body is the solution of differential equations~(\ref{eq:1-4}), that can be obtained by any ODE solver for given external forces and initial state (pose and velocities). 

\begin{figure}[t]
\centering
\includegraphics[width=0.7\columnwidth]{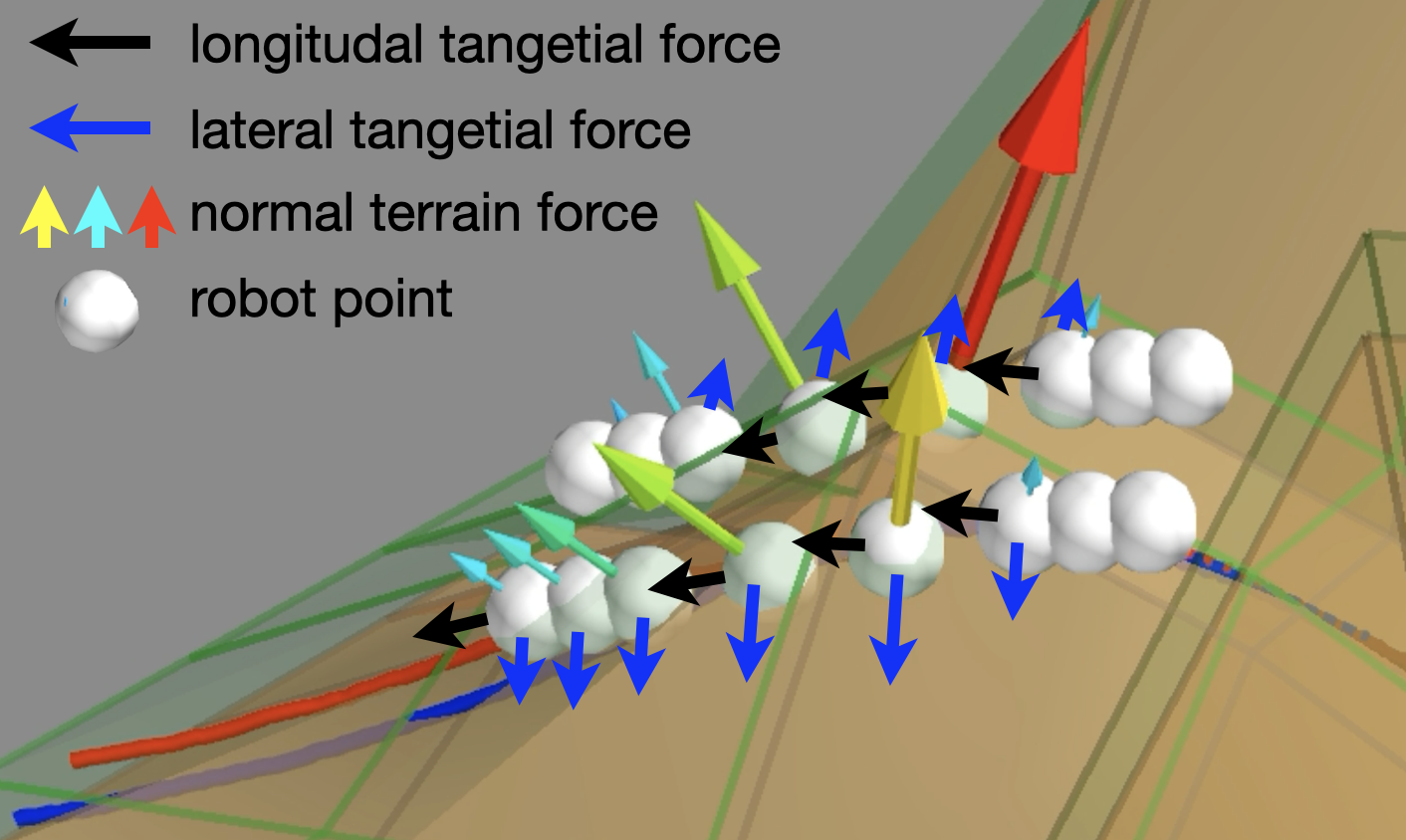}
\caption{\textbf{Three types of forces} acting on each robot point which is in contact with the terrain. Only track points are visualized for the bravity.
  }
  \label{fig:three_forces}
\end{figure}

When the robot is moving over a terrain, two types of external forces are acting on the point cloud $\mathcal{P}$ representing its model: (i) gravitational forces and (ii) robot-terrain interaction forces. The former is defined as $\mathbf{f}_{gi} = [0, 0, -m_ig]^\top$ and acts on all the points of the robot at all times, while the latter is the result of complex physical interactions that are not easy to model explicitly and act only on the points of the robot that are in contact with the terrain. There are three types of robot-terrain interaction forces: (i) normal terrain force that prevents the penetration of the terrain by the robot points, (ii) longitudinal tangetial force that generates forward acceleration when the tracks are moving, and (iii) the lateral tangetial force that prevents side slippage of the robot; see Figure~\ref{fig:three_forces} for details.

\subsection{Normal reaction forces}~\label{subsec:normal_forces}
One extreme option is to predict the 3D force vectors $\mathbf{f}_i$ directly by the convolutional network, but we decided to enforce additional prior assumptions to reduce the risk of overfitting. These prior assumptions are based on common intuition from the contact dynamics of flexible objects. In particular, we assume that the magnitude of the force that the terrain exerts on the point $\mathbf{p}_i\in \mathcal{P}$ increases proportionally to the deformation of the terrain. Consequently, the network does not directly predict the force, but rather predicts the height of the terrain $h\in\mathbb{R}$ at which the force begins to act on the robot body and the stiffness of the terrain $e\in\mathbb{R}^+$. We understand the quantity $e$ as an equivalent of the spring constant from Hooke's spring model. Given the stiffness of the terrain and the point of the robot that penetrated the terrain by $\Delta h$, the reaction force is calculated $e\cdot \Delta h$. 

Since such a force, without any additional damping, would lead to an ethernal bumping of the robot on the terrain, we also introduce a robot-terrain damping coefficient $d\in\mathbb{R}^+$, which similarly reduces the force proportionally to the velocity of the point that is in contact with the terrain.

\textbf{Terrain model:} In order to apply these forces, we train the terrain encoder to predict a $4$-channel heightmap that contains these quantities (height, stiffness, damping and friction coefficient). The heightmap is in the robot's coordinate frame from the time the input camera image has been captured. During the solution of differential equations~(\ref{eq:1-4}), we apply forces $\mathbf{f}_i$ computed from the terrain that is in contact with point $\mathbf{p}_i = [\mathbf{p}_{xi}, \mathbf{p}_{yi}, \mathbf{p}_{zi}]^T$. We will always denote the quantities corresponding to point $\mathbf{p}_i$ as $h_i, e_i, d_i$. We considered two possible terrain models: (i) piece-wise constant model, in which the corresponding quantity is uniquely determined by rounding on the heightmap grid and (ii) bilinear model, in which the quantities are estimated by bilinear interpolation from neighboring bins. Since the former exhibited significantly lower stability, less realistic behavior, and diminishing gradients during backpropagation, we stayed with the latter; see Figure~\ref{fig:optim} for an example of the bilinear terrain model.

\begin{figure}[t]
\centering
\includegraphics[width=0.4\textwidth]{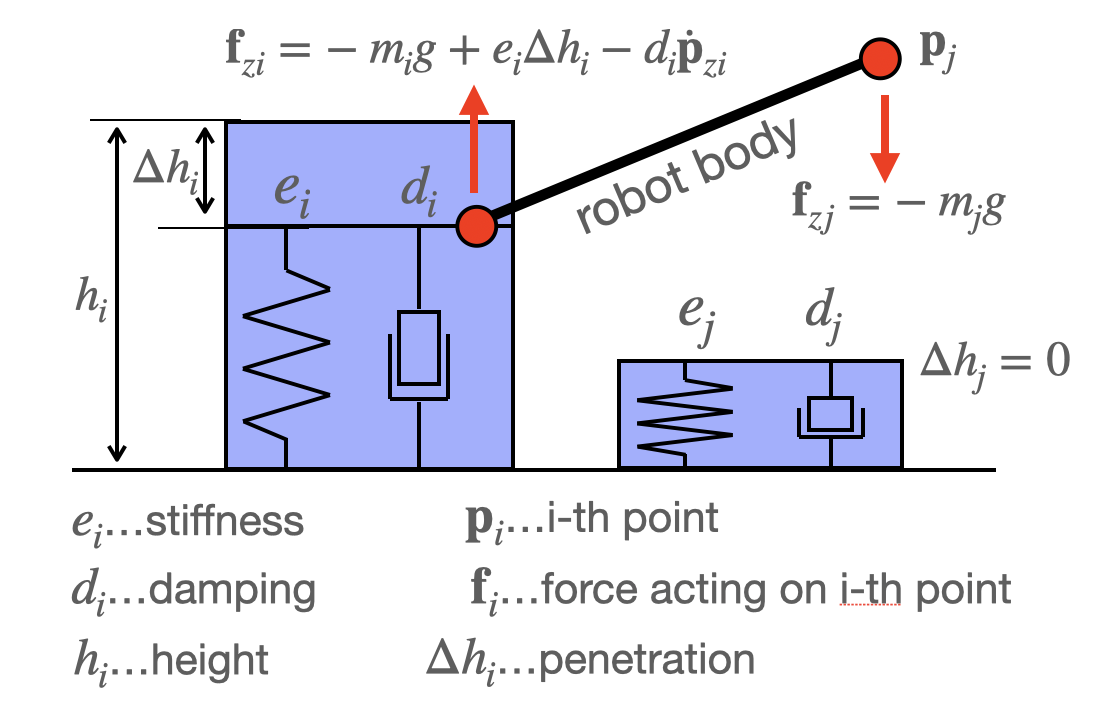}
  \caption{\textbf{Vertical terrain force model:} Simplified 2D sketch demonstrating vertical forces acting on a~robot body consisting of two points $\mathbf{p}_i$ and $\mathbf{p}_j$. 
  }
  \label{fig:robot-terrain_forces}
\end{figure}

\textbf{Vertical terrain force model:} The most naive application of the principles outlined above assumes that the terrain induces only vertical forces. The vertical force $\mathbf{f}_{zi}$ acting on the $i$-th point $\mathbf{p}_i$ of the robot body is then computed as follows:
$$
\mathbf{f}_{zi} = -m_ig + \begin{cases} 
 e_i(h_i-\mathbf{p}_{zi}) - d_i\dot{\mathbf{p}}_{zi}  & \text{if } \mathbf{p}_{zi}\leq h_i \\
0 & \text{if } \mathbf{p}_{zi}> h_i      %
\end{cases}
$$
See Figure~\ref{fig:robot-terrain_forces} for a visual sketch of terrain properties and forces on a simplified 2-point robot body. These vertical forces can influence only the robot's roll, pitch, and height; therefore, its initial velocity in ($x$,$y$, heading)-space remains constant in the ODE solution. 

\textbf{Normal terrain force model:} The more advanced model applies similar forces in the normal direction $\mathbf{n}_i$ of the terrain surface, where the $i$-th point is in contact with the terrain.
$$
\mathbf{f}_{i} = [0,0,-m_ig] + \begin{cases} 
 (e_i\Delta h_i - d_i(\dot{\mathbf{p}}_{i}^\top\mathbf{n}_i))\mathbf{n}_i  & \text{if } \mathbf{p}_{zi}\leq h_i \\
\mathbf{0} & \text{if } \mathbf{p}_{zi}> h_i      %
\end{cases},
$$
where terrain penetration $\Delta h_i = (h_i-\mathbf{p}_{zi})\mathbf{n}_{zi}$ is estimated by projecting the vertical distance on the normal direction.

\subsection{Tangential (driving) forces}~\label{subsec:tangential_forces}
Our~tracked robots move by moving the main tracks and flippers (auxiliary tracks). The flipper motion is purely kinematic in our model. This means that in a given time instant, their pose is uniquely determined by a $4$-dimensional vector of their rotations, and they are treated as a rigid part of the robot. 

The motion of the main tracks is transformed into forces tangential to the terrain. In particular, if the point on the main track is in contact with the terrain, two tangential forces are generated: longitudinal force $f_{long}$, which is parallel with the current orientation of the robot, and lateral force $f_{lat}$, which is perpendicular to the current orientation of the robot.

\textbf{The longitudinal force} delivers forward acceleration of the robot when robot tracks (either on flippers or on main tracks) are moving. When robot point $\mathbf{p}_i$, which belongs to a track, is in contact with terrain with friction coefficient $\mu_i$, the magnitude of the resulting longitudinal force is computed as follows:
$f_{long, i} = \mu_i m_ig\big(\sigma(u-v_x)-0.5\big),$
where $u$ is the velocity of the track, and $v=[v_x,v_y]$ is the velocity of the point $\mathbf{p}_i$ with respect to the terrain transformed into the robot coordinate frame. Consequently, the higher the difference $u-v_x$, the higher the longitudinal force. The maximum of this force is given by the gravitation force and the friction coefficient.

\textbf{The lateral force} avoids the side slippage of the robot on the terrain. When robot point $\mathbf{p}_i$ is in contact with terrain, the magnitude of the resulting lateral force is computed as follows:
$f_{lat, i} = \mu_i m_ig\big(\sigma(-v_y)-0.5\big).$
Similarly, the higher the side velocity $v_y$, the higher the force that prevents this motion. This model can be understood as a simplified Pacejka's tire-road model~\cite{pacejka-book-2012} that is popular for modeling tire-road interactions.

\subsection{Implementation}

\textbf{Differentiable implementation of ODE solver:} The overall learning behavior (speed and convergence) is mainly determined by (i) the way the differentiable ODE solver is implemented and (ii) the horizon over which the difference between trajectories is optimized (i.e., the distance of control waypoints from the initial state).
We denote current state of the robot as $\mathbf{s} = [\mathbf{x}, \mathbf{v}, \mathbf{R}, \boldsymbol{\omega}]$ and waypoint $\mathbf{u}$ representing the control signal. We implemented the forward model $\dot{\mathbf{s}}= g(\mathbf{s}, \mathbf{u})$ corresponding to differential equations~(\ref{eq:1-4}) in PyTorch.  
We experimented with two possible implementations of the ODE solver. The first one was based on the Neural ODE framework~\cite{neural-ode-2021}. This implementation leverages the existing ODE solver available in Scipy. The backward pass is computed by calling the solver on the adjoint ODE constructed through PyTorch's autodiff functionality. The second implementation directly implements the Euler integration method with a fixed step size in PyTorch that builds the full computational graph of the forward integration and then uses the autograd functionality to get the gradient. Since the former turned out to be significantly faster and achieved comparable accuracy, we employed it for most of our experiments.

\subsection{Learning and losses}~\label{subsec:learning_and_lossses}
\textbf{Trajectory loss:} Since the architecture (Figure~\ref{fig:architecture}) is end-to-end differentiable, we can directly learn to predict all these intermediate outputs just from trajectory error 
\begin{equation}~\label{eq:traj_loss}
    \mathcal{L}_\tau = \|\tau-\tau^\star\|^2,
\end{equation}
where $\tau^\star$ is the ground truth trajectory obtained by a common SLAM procedure from onboard measurements only. In order to make the implementation more efficient and explainable, we employ several regularization losses. 

\begin{figure}[t]
\centering
\includegraphics[width=\columnwidth]{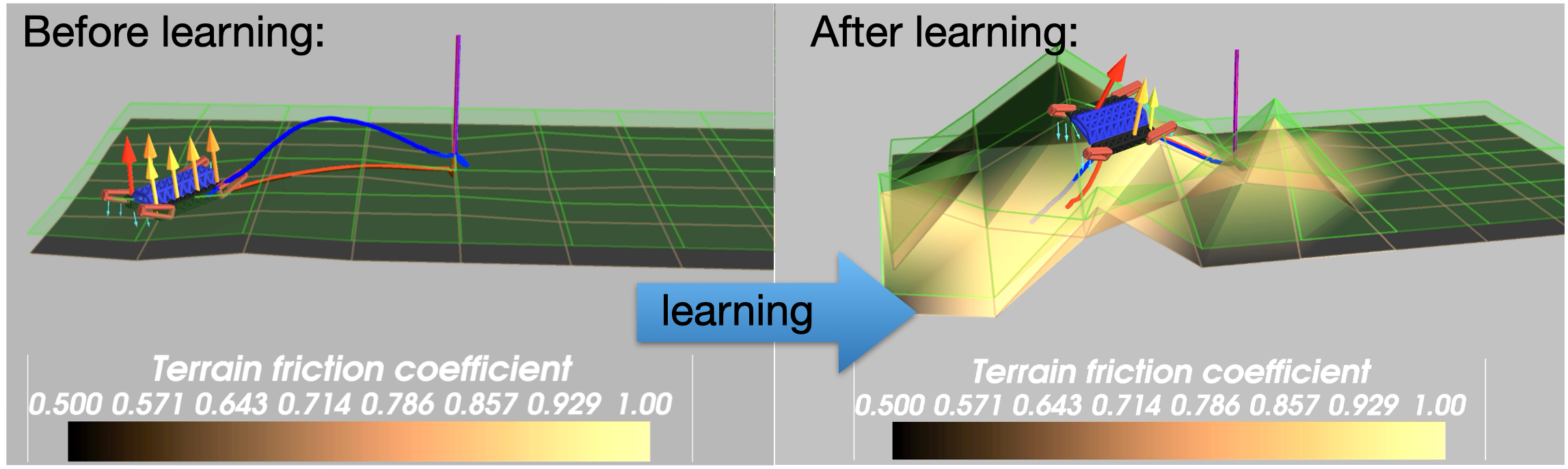}
\caption{\textbf{Terrain shape computed by backpropagating through $\nabla$Physics:} Shape of the terrain (border of the area where terrain forces start to act) outlined by mesh and interaction forces outlined by different colors. The optimized trajectory is in red, and the ground truth trajectory is in blue.}
\label{fig:optim}
\end{figure}

\textbf{Terrain loss:} Since backpropagating through the physics engine is time-consuming, we first precompute terrain heightmaps $\mathcal{H}_t^\star$ by backpropagating from each ground truth trajectory toward terrain properties, see Figure~\ref{fig:optim} for an example. These terrain heightmaps serve as an auxiliary ground truth that prevents costly backpropagation through the physics engine in every iteration or if any learning hyper-parameters are changed. Consequently, we introduce terrain loss
\begin{equation}~\label{eq:terrain_loss}
    \mathcal{L}_t = \|\mathbf{W}_t\circ(\mathcal{H}_t-\mathcal{H}_t^\star)\|^2 + \lambda_d\|\mathcal{H}_d\|^2,
\end{equation}
where $\mathbf{W}_t$ denotes the array, which masks out heightmap cells untouched by the ground truth trajectory, and $\circ$ is the element-wise product. The term $\|\mathcal{H}_d\|^2$ just provides additional regularization by providing a penalty for decreasing the geometrical heightmap.

\textbf{Geometrical loss:} In order to further regularize the learning procedure, we leverage existing lidar measurements in our dataset and construct ground truth geometrical heightmap $\mathcal{H}_g^\star$ from lidar point clouds using the same method as in \cite{wang2011new}. Consequently, we introduce geometrical loss
\begin{equation}~\label{eq:geom_loss}
    \mathcal{L}_g = \|\mathbf{W}_g\circ(\mathcal{H}_g-\mathcal{H}_g^\star)\|^2,
\end{equation}
where $\mathbf{W}_g$ denotes the array, which masks out heightmap cells measured by the lidar scan.

\begin{figure*}
{\includegraphics[width=\textwidth]{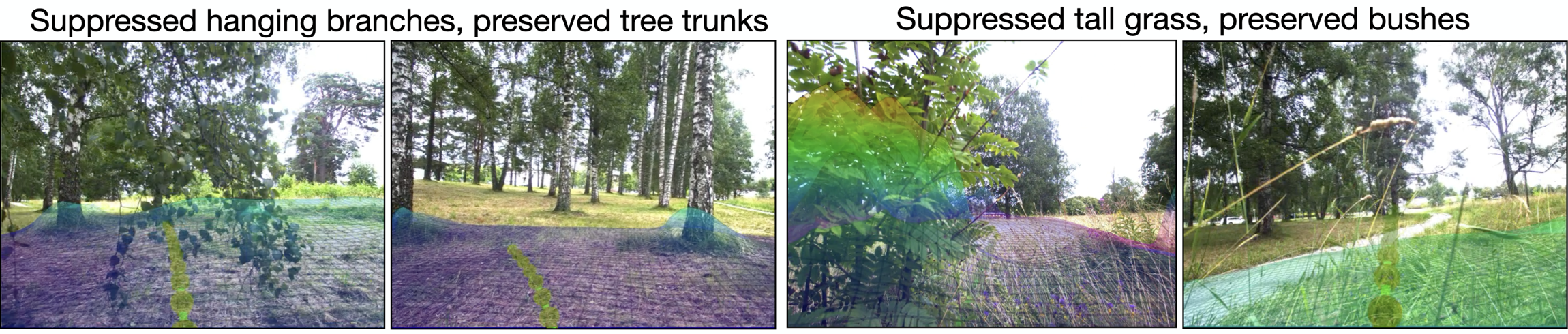}}
\caption{\textbf{Qualitative results}: Predicted terrain heightmap and robot trajectory given given camera images. The terrain heightmap corresponds to the expected heights at which the terrain begins to generate counter-penetration forces. Maximum height limited to 1m for practical reasons.}\label{fig:qualitative_results}
\end{figure*}

For the terrain encoder training, we do not require any tedious manual annotations.
The geometric heightmap, $\mathcal{H}_{g}$, is directly estimated from a lidar scan, while only a recorded (with the help of SLAM or GPS) robot trajectory $\tau$ is required to obtain the terrain properties $\mathcal{H}_{t}$ term.
Therefore, the terrain encoder training process is fully self-supervised.

\textbf{Prediction horizon:} The length of the horizon employed in learning should correspond to the horizon at which the method is intended to be used. Nevertheless, we observed that both of the previous implementations suffered from diminishing gradients when the horizon was too long. Consequently, we split the training trajectories into 1-second--long chunks.
Once trained, this function can be directly usable as a node expansion model in any state-of-the-art planner,  MDP controller or terrain traversability predictor (e.g. by thresholding the maximum roll and pitch of the robot in the resulting trajectory). Figure~\ref{fig:qualitative_results} shows qualitative examples of the predicted terrain and robot trajectories.

\section{EXPERIMENTS}
\subsection{Data}

\begin{figure}
    \centering
    
    \subfigure[\textit{Wheeled} Husky robot]{\includegraphics[width=0.38\columnwidth]{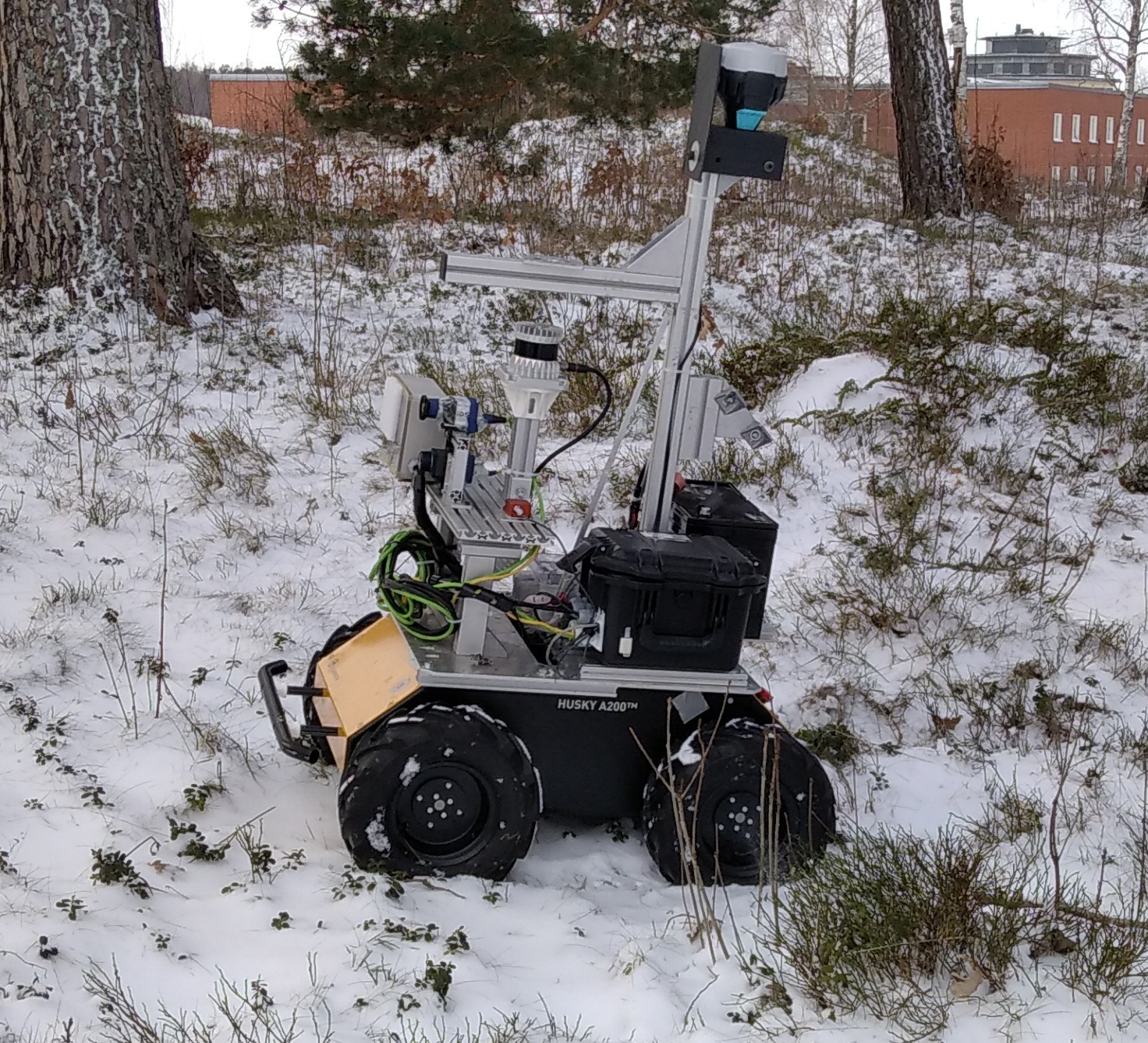}}
    \subfigure[\textit{Tracked} MARV robot]{\includegraphics[width=0.44\columnwidth]{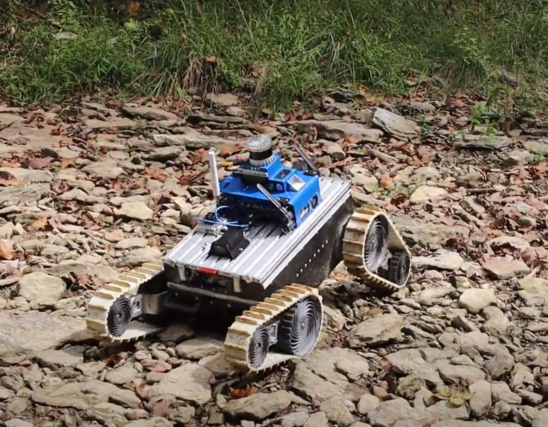}}
    
    \caption{Robot platforms used to collect RobinGas data sequences.}
    \label{fig:robot_platforms}
\end{figure}
We pretrain the terrain encoder on the large-scale outdoor dataset RELLIS-3D~\cite{jiang2020rellis3d}.
It is a multimodal dataset collected in an off-road environment containing accurately localized (with RTK GPS) $13,556$ lidar scans and corresponding time-synchronized RGB images.
Despite the amount and annotation quality of the data provided in the RELLIS-3D sequences, it lacks examples of a robot moving over hills, obstacles, and traversing high grass.
To fill this gap in autonomous off-road navigation, we release new data sequences, called RobinGas, containing sensory data from forest and field scenarios recorded with the two robot platforms: \textit{tracked} (MARV robot), \textit{wheeled} (Clearpath Husky robot), Figure~\ref{fig:robot_platforms}.
The data set contains point cloud scans from Ouster OS0-128, OS0-32 lidars and corresponding RGB images from Basler and IDS cameras installed on the robots (front, rear, left, and right).
For each lidar-images data sample, we additionally record robot localization data that was obtained using ICP SLAM~\cite{Pomerleau-2013-AR}. It showed to be a better choice than GPS that is unreliable near or under forest canopy.
With the help of the $\nabla$physics module, we obtain height values for each cell of the terrain traversed by the robot following a~\qty{10}{\second}--long trajectory. The height values are being optimized in such a way that the trajectory predicted by the $\nabla$physics module is as close as possible to the SLAM-generated ground-truth, as described in Section~\ref{subsec:terrain_encoder}.

After training the MonoForce terrain encoder and the LSS~\cite{philion2020lift} model on RELLIS-3D dataset, we fine-tune them on our RobinGas data sequences.
For evaluation, we select the best-performing model on validation part of the RobinGas data in terms of criterion:
$\mathcal{L} = \mathcal{L}_t + \lambda_g \mathcal{L}_g,$
where $\mathcal{L}_t$ and $\mathcal{L}_g$ are defined by equations (\ref{eq:terrain_loss}), (\ref{eq:geom_loss}), $\lambda_g$ is a hyperparameter.

\subsection{Trajectory Estimation Accuracy}~\label{subsec:traj_est_acc}

Experiments are conducted according to the pipeline represented in Figure~\ref{fig:overview}.
The following terrain encoding methods are compared:
\begin{itemize}
    \item heightmap interpolation from a lidar scan~\cite{wang2011new},
    \item model~\cite{Salansky-RAL} providing geometrical heightmap $\mathcal{H}_{g}$ from a lidar scan,
    \item LSS model~\cite{philion2020lift} providing $\mathcal{H}_{g}$ from RGB images,
    \item MonoForce (\textbf{ours}) predictor providing terrain properties $\mathcal{H}_{t}$ from RGB images.
\end{itemize}

\begin{figure*}
    \newcommand\imgwidth{0.5}
    \centering
    

    \subfigure[\textbf{$\mathbfcal{H}_{\mathbf g}$ by~LSS~\cite{philion2020lift}}: Predicted geometrical heightmap, predicted trajectory, terrain interaction forces.]{\includegraphics[width=\imgwidth\columnwidth]{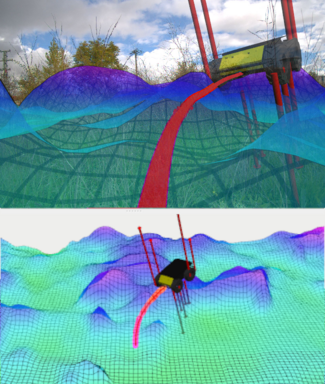}}
    \subfigure[$\mathbfcal{H}_{\mathbf{d}}$: Heightmap difference predicted by (MonoForce, \textbf{ours}).]{\includegraphics[width=\imgwidth\columnwidth]{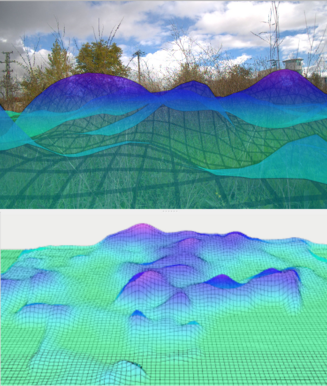}}
    \subfigure[$\mathbfcal{H}_{\mathbf t}=\mathcal{H}_{g}-\mathcal{H}_{d}$: Predicted supporting terrain, predicted trajectory, terrain interaction forces (MonoForce, \textbf{ours}).]{\includegraphics[width=\imgwidth\columnwidth]{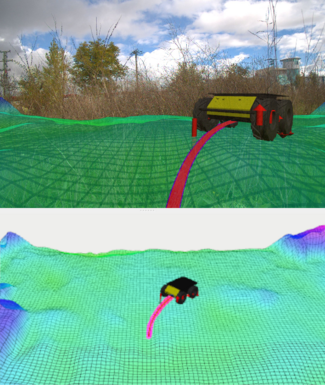}}
    \subfigure[\textbf{$\mathbfcal{H}_{\mathbf g}$ by~\cite{wang2011new}}: Interpolated lidar heightmap, predicted trajectory on the heightmap.]{\includegraphics[width=\imgwidth\columnwidth]{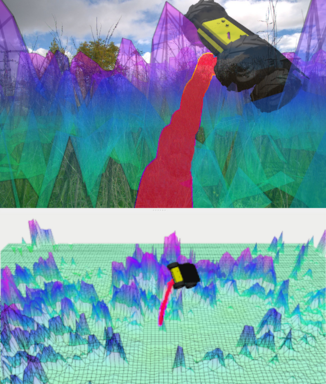}}
    
    \caption{\textbf{Qualitative comparison of different heightmap estimation methods:} $\mathcal{H}_{g}$ estimation with LSS model~\cite{philion2020lift}, our MonoForce terrain encoder, and $\mathcal{H}_{g}$ interpolation from lidar cloud~\cite{wang2011new}. Four RGB images are given as input to LSS and MonoForce. \textit{Top row}: front camera image with projected heightmaps and trajectories. \textit{Bottom row}: estimated heightmaps in 3D, predicted trajectories for given control input, and normal component of robot-terrain interaction forces, red arrows at contact points. The length of the arrows denotes the force magnitude. In all scenarios, the Husky robot starts from a stationary state and is given constant linear and angular velocity inputs $\mathbf{u} = \{ (\qty{0.5}{\meter\per\sec}, \qty{0.13}{\radian\per\sec}) \}_{t}$, $t\in [0..\qty{7}{\sec}]$.}
    
    \label{fig:hm_predictions}
\end{figure*}

Given the $\mathcal{H}_{t}$, robot model, and a sequence of control commands driving the robot, the $\nabla$physics module outputs the predicted trajectory $\tau = \{(\textbf{x}_i, \textbf{R}_i)\}_i$.
This trajectory is then compared to $\tau^{*} = \{(\textbf{x}^{*}_i, \textbf{R}^{*}_i)\}_i$, the ground-truth trajectory computed by the SLAM pipeline~\cite{Pomerleau-2013-AR}.
We compute mean translation and orientation errors $\Delta\mathbf{x}, \Delta\mathbf{R}$ for robot locations in the time moments corresponding to ground truth poses $i\in[1\dots{T}]$.
The metrics are computed for the trajectory parts that are inside the predicted heightmap region.
\begin{equation}
    \Delta\mathbf{x}=\frac{1}{N}\sum_{i=1}^T \|\mathbf{x}_i - \mathbf{x}_i^{*}\|
\end{equation}
\begin{equation}
    \Delta\mathbf{R}=\frac{1}{N}\sum_{i=1}^{T}\arccos\frac{\trace({\mathbf{R}^{\top}_i\mathbf{R}^{*}_i})-1}{2}
\end{equation}

Table~\ref{tab:traj_accuracy_rellis3d} contains the results for the test part of RELLIS-3D data.
It could be noticed that the LSS~\cite{philion2020lift} and MonoForce methods are on-par with the methods predicting geometrical heightmap from lidar scans in terms of position estimation metric, $\Delta\mathbf{x}$.
It could be explained by the fact that the RELLIS-3D environments do not have many examples of the robot driving through flexible terrain.
So the actual robot footprint supporting terrain does not differ much from its geometrical shape.
The robot's orientation estimation w.r.t. supporting terrain remains challenging ($\Delta\mathbf{R}$ column in the Table~\ref{tab:traj_accuracy_rellis3d}) for the image-based model in comparison to the geometrical ones.

On the contrary, the results for the RobinGas data indicated in the Table~\ref{tab:traj_accuracy} are more interesting.
Both image-based approaches (\cite{philion2020lift} and MonoForce) outperform the geometrical methods (\cite{wang2011new}, \cite{Salansky-RAL}) for terrain shape estimation on RobinGas data.
\begin{table}
    \caption{Trajectory estimation accuracy on RELLIS-3D data.}\label{tab:traj_accuracy_rellis3d}
    \centering
    \begin{tabular}{c | c | c | c | c}
    \hline
    input & terrain encoder & $\tau$ pred. & $\Delta \mathbf{x}$ [\si{\meter}] & $\Delta\mathbf{R}$ [\si{\deg}] \\
    \hline

    point cloud & $\mathcal{H}_{g}$ interp. \cite{wang2011new} & $\nabla$physics & 0.14 & \textbf{4.06}  \\
    \hline

    point cloud & $\mathcal{H}_{g}$ pred. \cite{Salansky-RAL} & $\nabla$physics & 0.20 & 4.63 \\
    \hline
    
    RGB images & $\mathcal{H}_{g}$, LSS~\cite{philion2020lift} & $\nabla$physics & \textbf{0.12} & 4.59  \\
    \hline

    RGB images & $\mathcal{H}_{t}$, MonoForce & $\nabla$physics & 0.13 & 4.53 \\
    \hline
    
    \end{tabular}
\end{table}
\begin{table}
    \caption{Trajectory estimation accuracy on RobinGas data.}\label{tab:traj_accuracy}
    \centering
    \begin{tabular}{c | c | c | c | c}
    \hline
    input & terrain encoder & $\tau$ pred. & $\Delta \mathbf{x}$ [\si{\meter}] & $\Delta\mathbf{R}$ [\si{\deg}] \\
    \hline

    point cloud & $\mathcal{H}_{g}$ interp. \cite{wang2011new} & $\nabla$physics & 0.13 & 4.74 \\
    \hline

    point cloud & $\mathcal{H}_{g}$ pred. \cite{Salansky-RAL} & $\nabla$physics & 0.13 & 4.32 \\
    \hline
    
    RGB images & $\mathcal{H}_{g}$, LSS~\cite{philion2020lift} & $\nabla$physics & 0.11 & 5.80  \\
    \hline

    RGB images & $\mathcal{H}_{t}$, MonoForce & $\nabla$physics & \textbf{0.07} & \textbf{2.62}  \\
    \hline
    
    \end{tabular}
\end{table}
Qualitative results are shown in Figure~\ref{fig:hm_predictions}. They provide comparison between our method output and the two geometrical heightmap estimation approaches: the LSS~\cite{philion2020lift} model taking as input RGB images and heightmap interpolation from lidar scan~\cite{wang2011new}.
As seen in the figure, our (MonoForce) system is able to correctly predict supporting terrain for a robot moving through high grass.
On the other hand, it is often not possible to predict feasible trajectories using the geometry-only methods, Figure~\ref{fig:hm_predictions}(d).

\subsection{Ablation Study}

\begin{figure}
    \centering    
    
    \subfigure[LSS model trained without geometrical loss (Table~\ref{tab:ablation}, 3-rd row)]{\includegraphics[width=0.55\columnwidth]{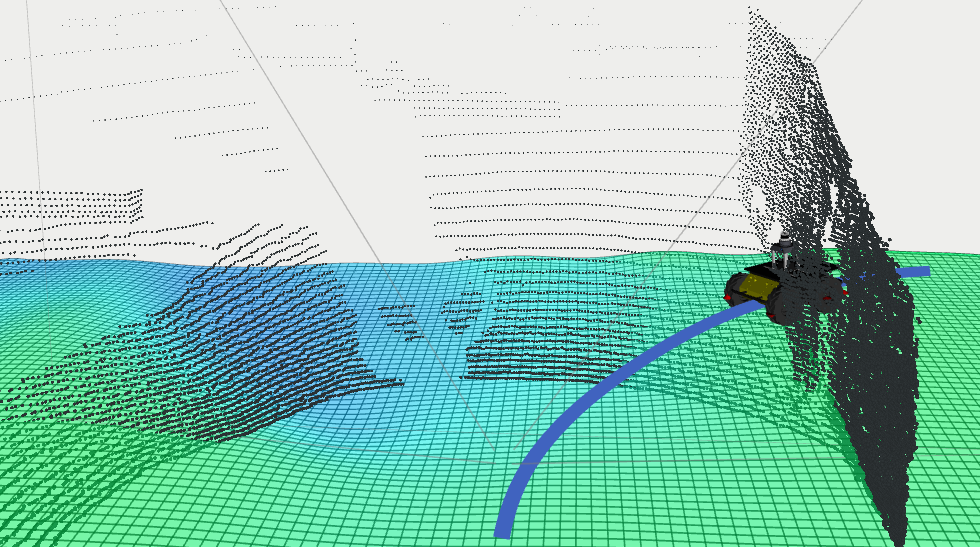}}
    \subfigure[MonoForce result (Table~\ref{tab:ablation}, 1-st row)]{\includegraphics[width=0.42\columnwidth]{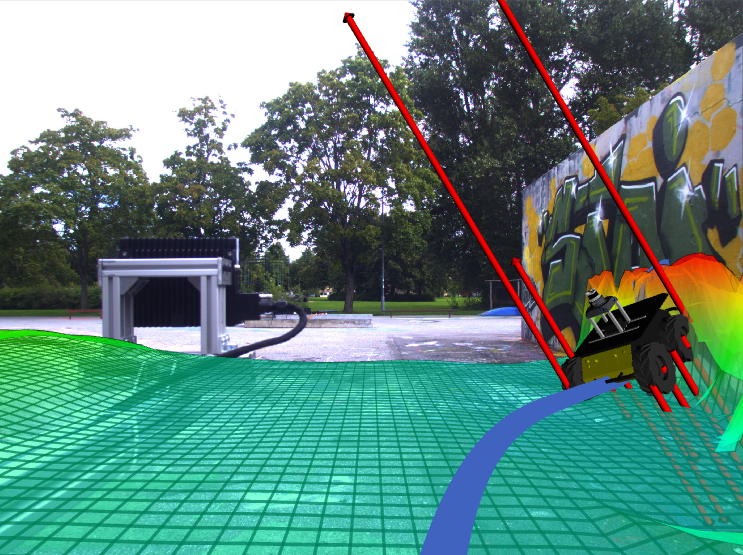}}
    
    \caption{Out-of-training distribution (robot-endangering) trajectory prediction. \textit{Left}: model trained without geometrical loss is not able to represent a wall in the heightmap. \textit{Right}: MonoForce model predicts the crash correctly.}
    
    \label{fig:ablation}
\end{figure}

We conduct the ablation studies to investigate an impact of individual components of the MonoForce pipeline (introduced in Figure~\ref{fig:architecture}) on the trajectory prediction accuracy.
The results are presented in the Table~\ref{tab:ablation}.
The evaluation is done in the same way as described in the Sec.~\ref{subsec:traj_est_acc} on the RobinGas dataset.
The top row corresponds to the complete MonoForce architecture results.
The second row denotes trajectory accuracy degrades when the terrain properties estimation step is switched-off (or only geometrical $\mathcal{H}_g$ heightmap is used to predict trajectories).
It is not obvious though from the third row results whether providing the model the knowledge about heightmap geometry helps to predict more accurate trajectories.
However, a model trained without geometrical information struggles to generalizes to out-of-distribution situations not present in the dataset, for example a robot crashing into rigid obstacles, Fig.~\ref{fig:ablation}.

\begin{table}
    \caption{Impact of MonoForce Components on Trajectory Accuracy}
    \centering
    \begin{tabular}{c | c | c}
    \hline
    
    method & $\Delta \mathbf{x}$ [\si{\meter}] & $\Delta\mathbf{R}$ [\si{\deg}] \\
    \hline
    
    Geom. $\mathcal{H}$ est. + Terrain est. +  $\nabla$physics & \textbf{0.07} & 2.62 \\
    \hline

    Geom. $\mathcal{H}$ est. +  $\nabla$physics & 0.11 & 5.80 \\
    \hline

    Terrain est. +  $\nabla$physics & 0.09 & \textbf{2.56} \\
    \hline

    
    \end{tabular}
    \label{tab:ablation}
\end{table}

\section{CONCLUSION and FUTURE WORK}

In this work, we have presented \emph{MonoForce}, an explainable, physics-aware and end-to-end differentiable model that predicts robot's trajectory from monocular camera images.
As a \emph{grey-box} model, it benefits from the end-to-end trainability in different domains while it still retains the determinism of its physics engine and explainability of its estimates of terrain properties.
The training process is self-supervised; it only requires the monocular camera images, lidar scans for evaluating the geometrical loss and reference trajectories for the terrain loss.
The model learns to recognize non-rigid obstacles similar to those that have been driven over in the reference trajectories, while keeping the understanding and prediction capabilities for obstacles it has not encountered yet.
It treats them as rigid obstacles until future data prove otherwise.

In the experiments, we have shown that MonoForce generates accurate geometric and terrain heightmaps that in turn serve as a basis for robot-terrain interaction force and trajectory estimates.
These estimates are valid both on rigid and non-rigid terrains.
On the rigid terrain that is present in RELLIS-3D dataset, our results are comparable to the lidar-based methods and to the LSS model.
The strength of MonoForce becomes apparent on non-rigid terrains, where we outperform the other methods with average \SI{0.07}{\meter} translation and \SI{2.62}{\degree} rotation error in the predicted robot trajectories.

One of the directions for future work is seen in designing different trajectory patterns in the data collection phase that would better cover the estimated terrain heightmap and thus result in more informative terrain loss.
We also plan to generalize the definition of supporting terrain to be defined by thresholding the predicted reaction force, which should allow representing a richer variety of terrain properties. It is also possible to define multiple layers of the terrain, each representing an independent physical trait and generating its own reaction force.

   

\subsection*{Acknowledgments}
This work was co-funded by the Grant Agency of the CTU in Prague under Project SGS24/096/OHK3/2T/13, the European Union under the project Robotics and advanced industrial production (reg. no. CZ.02.01.01/00/22\_008/0004590), the Czech Science Foundation under Project 24-12360S, and the European Union's Horizon Europe Framework Programme under the RaCOON project (ID: 101106906).

\bibliographystyle{IEEEtran}
\bibliography{IEEEabrv, main}


\end{document}